\pdfoutput=1

\documentclass[11pt]{article}

\usepackage[]{acl}
\usepackage{times}
\usepackage{latexsym}
\usepackage{amsmath,amssymb}
 
\usepackage[T1]{fontenc}

\usepackage[utf8]{inputenc}

\usepackage{microtype}
\usepackage{graphicx}
\usepackage{subcaption}
\usepackage{booktabs}

\usepackage{float}
\usepackage{longtable}
\usepackage[export]{adjustbox}

\usepackage{makecell}

\usepackage{arydshln}

\title{Training Language Models with Language Feedback}

\author{Jérémy Scheurer$^{1}$ ~~ Jon Ander Campos$^{1~2}$ ~~Jun Shern Chan$^{1}$ ~~ \textbf{Angelica Chen}$^1$ \\ \textbf{Kyunghyun Cho}$^{1~3~4}$  ~~ \textbf{Ethan Perez}$^1$ \\
$^1$New York University, $^2$University of the Basque Country, $^3$Genentech, $^4$CIFAR LMB\\
  {\tt \{jeremy.scheurer,perez\}@nyu.edu} \\}

\begin{document}
\maketitle
\begin{abstract}
Pretrained language models often do not perform tasks in ways that are in line with our preferences, e.g., generating offensive text or factually incorrect summaries.
Recent work approaches the above issue by learning from a simple form of human evaluation: comparisons between pairs of model-generated task outputs.
Comparison feedback conveys limited information about human preferences per human evaluation.
Here, we propose to learn from natural language feedback, which conveys more information per human evaluation.
We learn from language feedback on model outputs using a three-step learning algorithm.
First, we condition the language model on the initial output and feedback to generate many refinements.
Second, we choose the refinement with the highest similarity to the feedback.
Third, we finetune a language model to maximize the likelihood of the chosen refinement given the input.
In synthetic experiments, we first evaluate whether language models accurately incorporate feedback to produce refinements, finding that only large language models (175B parameters) do so. Using only 100 samples of human-written feedback, our learning algorithm finetunes a GPT-3 model to roughly human-level summarization ability.
\end{abstract}

\section{Introduction}

\begin{figure}[ht]
    \centering
    \includegraphics[width=.48\textwidth,keepaspectratio]{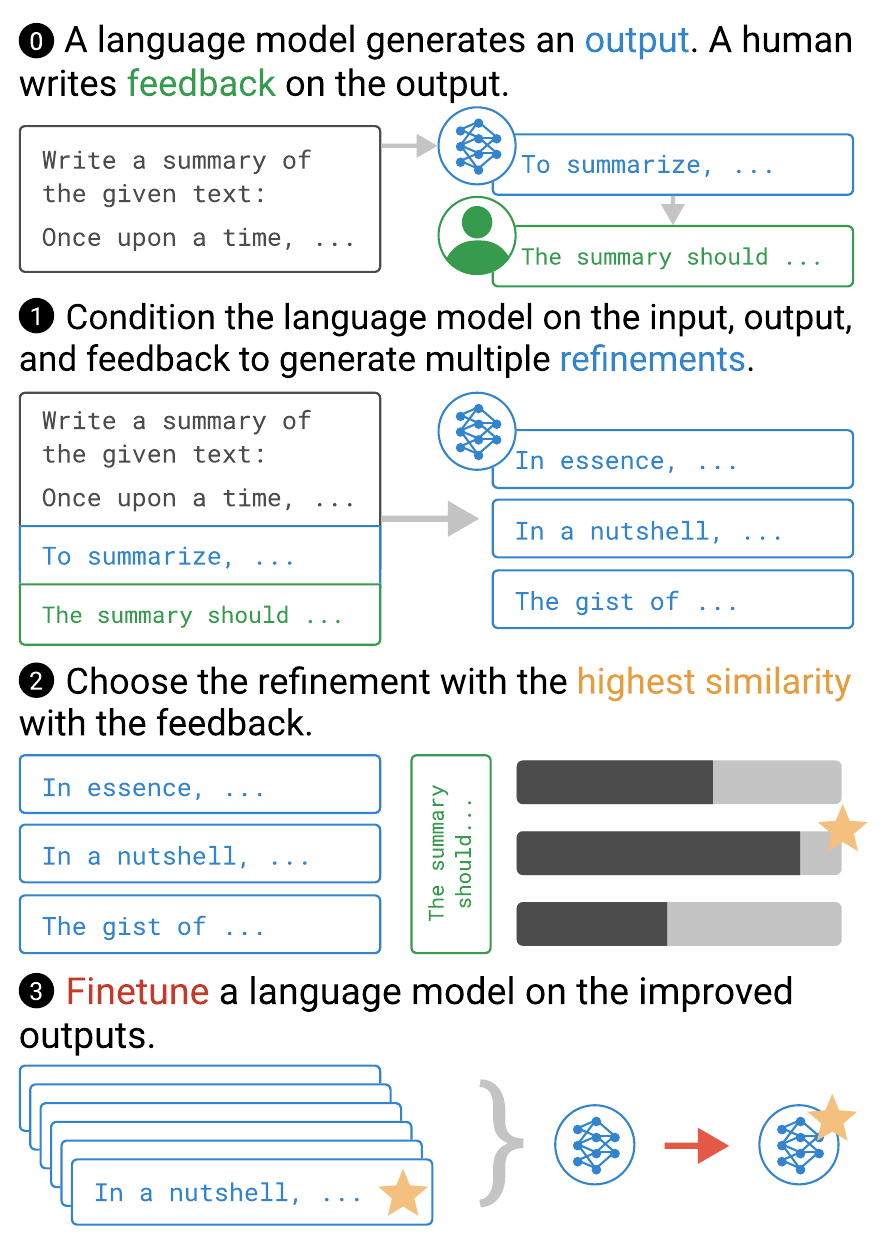}
    \caption{An overview of our algorithm for learning from natural language feedback.}
    \label{fig:illustration}
\end{figure}

Language Models (LMs) achieve strong performance across diverse NLP tasks, from summarization to question answering and conversational assistants \citep[][\textit{interalia}]{Radford2018ImprovingLU,Radford2019LanguageMA,brown2020language, rae2021scaling}. A key problem with LMs is that they generate text that violates human preferences, such as LM-generated misinformation \cite{lin2021truthfulqa}, offensive language \cite{gehman2020realtoxicityprompts}, and factually incorrect outputs such as summaries \cite{stiennon2020learning}.
Current methods alleviate such issues by training LMs to generate text that scores highly according to human preferences, or a predictive model thereof~\cite{ziegler2019fine,stiennon2020learning, nakano2021webgpt,ouyang2022training}. In this line of work, human evaluators indicate their preferences by comparing text outputs. However, each comparison provides little information per evaluation about human preferences.

We propose to use natural language feedback, which contains more information per evaluation.  We introduce a three-step learning algorithm, as shown in \autoref{fig:illustration}. First, we condition an LM on an input, model-generated output, and human-written feedback to sample many possible refinements of the output. Second, we choose the refinement with the highest embedding-based similarity with the feedback. Third, we finetune an LM on the chosen refinements. Our algorithm departs from prior work, which uses reinforcement learning methods~\citep[][\textit{inter alia}]{ziegler2019fine} or auxiliary losses~\cite{stacey2021natural} that cannot be straightforwardly generalized to using natural language feedback.

We validate our algorithm on a carefully-controlled synthetic task of removing offensive words from a sentence with GPT-3-based models~\citep{brown2020language,ouyang2022training}.
We find that only the largest GPT-3-based models (175B parameters) accurately refine outputs.
Using the above insight, we use the largest GPT-3 models to test our algorithm on text summarization, following~\citet{stiennon2020learning}.
A model trained with our algorithm generates summaries that human evaluators prefer to human reference summaries $\sim$51\% of the time.
We obtain these results when learning from only $100$ samples of natural language feedback.
Our analysis shows that LM-generated refinements typically incorporate the feedback, especially when choosing the refinement with the highest similarity with the feedback.
Our results suggest that natural language feedback is a promising avenue for learning from human preferences.


\section{Method}
Here, we define our problem formulation more formally.
Given an input $x$, we seek to generate an output $y$ that is high quality according to human preference judgments.
We assume access to natural language feedback $f$ on an initial model-generated output $y$ given the input $x$.

To tackle the above problem, we leverage the ability of pretrained LMs to follow instructions~\cite{Radford2019LanguageMA,sanh2021multitask,wei2022finetuned,ouyang2022training}.
We assume access to an LM that takes an input (e.g., a task instruction) and produces a distribution over text completions (e.g., a task output).
We instruct the LM to refine the initial output $y$ given the input $x$ and feedback $f$.
We then sample $N$ refinements $y'_1, \dots, y'_{N}$ from the LM.
Refinements may vary in quality, so we introduce a function $S$ that scores refinements for how effectively they incorporate feedback.
We choose the refinement with the highest score from $S$ and finetune a model on all chosen $y'$ given $x$. We use the resulting model to generate outputs at test time.

\begin{table}[t!]
\resizebox{\columnwidth}{!}{
\begin{tabular}{c c c c c} \toprule
Models  & \makecell{Ada  \\ ($\sim 350$M)} & \makecell{Babbage \\ ($\sim 1.3$B)} & \makecell{Curie \\ ($\sim 6.7$B)} & \makecell{Davinci \\ ($175$B)}  \\
\hline
GPT-3 & $1.0 \pm 0.3 $ & $1.1 \pm 0.3$ & $8.7\pm 0.8$ & $38.5\pm 1.3$ \\
InstructGPT & $1.6 \pm 0.3$ & $2.5 \pm 0.4$ & $5.4 \pm 0.6 $ & $35.6 \pm 1.3$ \\

 \bottomrule
\end{tabular}}
\caption{We report the accuracy in $\%$ with the standard error. On the task of removing offensive words from a sentence, only large LMs incorporate feedback.}
\label{tab:targeted_word_removal}
\end{table}

\section{Experiments}
\subsection{Can Language Models Use Feedback?}
\label{sec:targeted_word_removal}
For our algorithm to work, LMs must be able to accurately incorporate feedback to generate refinements.
Thus, we first validate our algorithm on a carefully-controlled synthetic task of removing specific offensive words from a given sentence.
We examine how effective various models are at incorporating feedback, to determine what model to use for refining outputs.

\paragraph{Experimental Setup}
We instruct an LM to refine an automatically-generated sentence with $\leq 10$ offensive words by removing $\leq 3$ specific words (see Appendix \ref{sec:word_removal_example} for a detailed explanation and examples). We evaluate how often the generated refinement exactly matches the target sentence, which we also automatically generate. For our LMs, we use differently-sized GPT-3 models~\citep{brown2020language} and their finetuned, InstructGPT counterparts~\citep{ouyang2022training}.\footnote{Via the \href{https://beta.openai.com/}{OpenAI API}. OpenAI does not disclose the size of the provided models, so we use estimates from \href{https://blog.eleuther.ai/gpt3-model-sizes/}{Eleuther}.} We report all hyperparameters used in Appendix \ref{sec:hyperparameters}.
We report mean and std. error for all results in our work.

\paragraph{Results}
Table \ref{tab:targeted_word_removal} shows the results. We observe that only the largest GPT-3 and InstructGPT models (175B parameters) incorporate feedback a non-negligible amount of time. Using this insight, we only use the 175B parameter (Davinci) models in the rest of our experiments.

\subsection{Text Summarization}
\label{sec:summarization_experiments}

\subsubsection{Experimental Setup}

\paragraph{Generating Refinements}
We now evaluate our algorithm on the real-world task of text summarization.
We follow prior work on learning from human preferences~\citep{stiennon2020learning} and learn to summarize Reddit posts from~\citet{volske-etal-2017-tl}.
We take $100$ samples from the Reddit data subset used in~\citet{stiennon2020learning}.
We use InstructGPT (175B) to generate initial summaries and refinements, using the instructions in Appendix \ref{sec:prompts}.
We then write feedback $f$ on the initial summary $y$ given the Reddit post $x$, and we generate possible refinements $y'_1, \dots, y'_{20}$.

\paragraph{Scoring Refinements} We choose a refinement with a scoring function $S$ that scores refinements for how effectively they incorporate feedback.
For $S$, we use contrastive pre-trained text embedding function $\mathcal{E}$ \citep{neelakantan2022text} to embed the feedback $f$ and refinements $y'_1, \dots, y'_{20}$\footnote{We use \href{https://beta.openai.com/}{OpenAI's API} to access the embeddings.}.
We then choose the refinement with the highest cosine similarity score with the feedback.
We opted for high similarity with the feedback because feedback often describes what the ideal or improved text would look like.
We refer to refinements generated with the above algorithm as \textsc{Refinement with Feedback + Best of N}.

\paragraph{Finetuning} 
We finetune GPT-3~\citep[175B;][]{brown2020language}\footnote{InstructGPT cannot yet be finetuning via OpenAI's API.} on refinements generated by \textsc{Refinement with Feedback + Best of N}. We compare against finetuning on \textsc{Initial Summaries} generated with InstructGPT.
We also compare against summaries generated directly by InstructGPT and GPT-3 (175B). We use the same instructions as for \textsc{Initial Summaries} (in Appendix \ref{sec:prompts}) and provide the post and its title.

\paragraph{Evaluation}
We test on $100$ unseen Reddit posts from the same dataset and conduct human evaluations for all experiments.\footnote{We plan to conduct larger-scale human evaluations in the future, to confirm our initial findings.} Evaluators rank the summaries according to the rubric in Appendix \ref{sec:human_evals}, with ties allowed.
We show the win rate of an algorithm, counting ties as a half win, similar to Kendall rank correlation.\footnote{\href{https://tinyurl.com/ba9mh4cy}{Kendall Rank correlation}}. We refer to Appendix \ref{sec:human_evals} for a description of all human evaluation and feedback annotation procedures and Appendix \ref{sec:ranking} for more details about the ranking scheme.

\begin{figure}[ht!]
    \centering
    \includegraphics[height=5.18cm,valign=t]{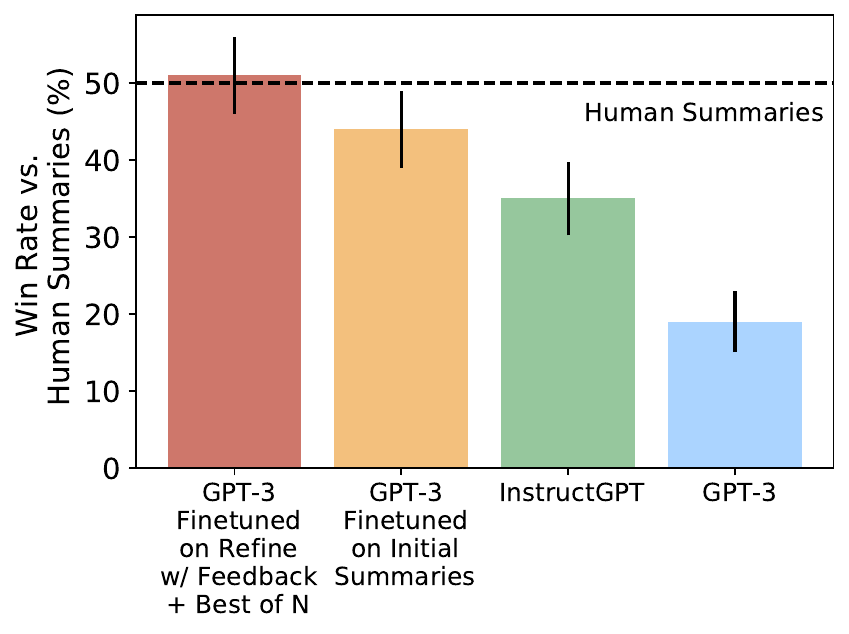}    
    \caption{How often human evaluators prefer summaries from our learning algorithm and baselines to
\textsc{Human Summaries}. Our proposed algorithm (leftmost bar) generates summaries of a similar quality to human summaries.}
    \label{fig:finetuning_win_rate}
\end{figure}

\subsubsection{Main Results}
Fig.~\ref{fig:finetuning_win_rate} reports the win rate of our learning algorithm over \textsc{Human summaries} and Appendix Fig.~\ref{fig:finetune_win_rate_instruct_gpt} reports the win rate over InstructGPT.
Finetuning on \textsc{Refinement with Feedback + Best of N} generates summaries on par with human summaries, with a win rate of $51.0 \pm 5.0 \%$ over human summaries.
In contrast, all baselines underperform human summaries, with win rates of $19.0 \pm 3.9\%$ (GPT-3), $35.0 \pm 4.8\%$ (InstructGPT), $44.0 \pm 5.0\%$ (finetuning on \textsc{Initial Summaries})).
In particular, our approach achieves a win rate of $57.0 \pm 5.0 \%$ over the strongest baseline, finetuning on \textsc{Initial Summaries}.
Our result suggests that our learning algorithm produces higher-quality summaries by finetuning on the higher-quality targets (our refinements).
Overall, we achieve strong results on summarization while learning from only $100$ samples of human-written feedback.
\begin{figure*}[t!]
\begin{minipage}[t]{.45\textwidth}
    \centering
    \includegraphics[scale=0.5,valign=t]{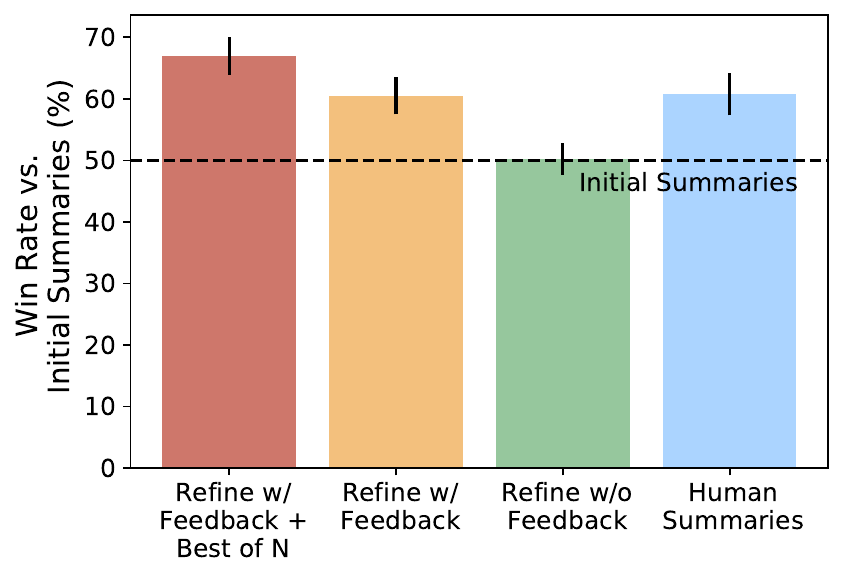}
\end{minipage}
\hfill
\begin{minipage}[t]{.45\textwidth}
     \centering
     \includegraphics[scale=0.5,valign=t]{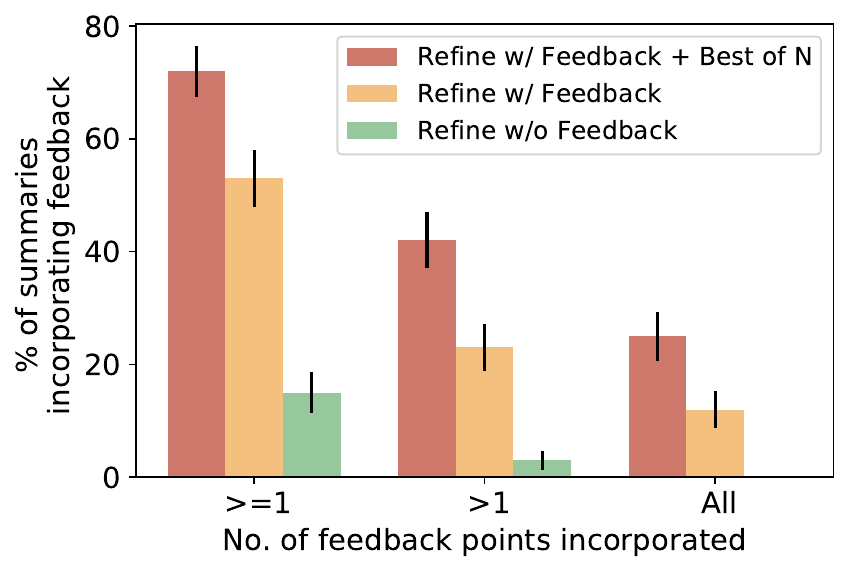}
\end{minipage}
\caption{\textbf{Left}: How often human evaluators prefer summaries from
each refinement method to the \textsc{Initial Summaries} (from InstructGPT). \textsc{Refinement with Feedback} improves on the \textsc{Initial Summaries} and outperforms human summaries with \textsc{Best of N} sampling. \textbf{Right}: Refining with feedback generally does incorporate specific point(s) mentioned in the feedback.}
\label{fig:win_rates_and_incorporating_feedback}
\end{figure*}
\subsubsection{Analysis}
We now aim to examine the importance of various aspects of our algorithm for generating high-quality refinements (before finetuning).
We evaluate \textsc{Refinement with Feedback}, which \textit{randomly} chooses a refinement $\in {y'_1, \dots, y'_{20}}$.
This ablation helps to evaluate the importance of choosing a refinement with our scoring function $S$.
We also evaluate \textsc{Refinement without Feedback}, which instructs the LM to refine the initial summary but without feedback.
This ablation helps to evaluate the importance of using the feedback.
Lastly, we evaluate \textsc{Human Summaries}, i.e., summaries written by Reddit users on their own posts, and \textsc{Initial Summaries}, i.e., the initial summary $y$ generated by the LM.
See Appendix \ref{sec:prompts} for concrete examples of the instructions that we use.

Fig.~\ref{fig:win_rates_and_incorporating_feedback} (left) shows the win rates of refinements from various methods against \textsc{Initial Summaries}. \textsc{Refinement with Feedback + Best of N} improves over the \textsc{Initial Summaries}, with our algorithm being preferred $67.0 \pm 3.1 \%$ of the time. Our algorithm is preferred $54.0 \pm 3.5 \%$ of the time to human summaries, while \textsc{Initial Summaries} are significantly worse than human summaries, preferred only $39.3\pm 3.4 \%$ of the time. Appendix Fig.~\ref{fig:win_rate_human_summary} shows win rates of refinements generated with various methods against \textsc{Human Summaries} and Appendix Fig.~\ref{fig:win_rate_across_original_summary_ranking} shows that refinements are more helpful when the initial summary is of lower quality. We also refer to Appendix \ref{sec:examples} for $10$ random examples of \textsc{Initial Summaries}, feedback, and refinements from various methods. Overall, using feedback and scoring refinements are both important steps for generating high-quality refinements of the initial output.

Here, we examine whether refinements are of higher quality because they incorporate the feedback, rather than by improving the summary in other ways.
To do so, we evaluate how often the refinements incorporate the human-written feedback.
We evaluate (1) how often $\geq 1$ point mentioned in the feedback is incorporated in the refinement, (2) how often $>1$ point is incorporated, and (3) how often all of the feedback is incorporated.
In Fig.~\ref{fig:win_rates_and_incorporating_feedback} (right), we see that our algorithm incorporates $\geq 1$ feedback point $72.0 \pm 4.5\%$ of the time, showing that LMs are able to incorporate feedback with high accuracy. \textsc{Refinements without Feedback} only incorporates at least one feedback point $15.0 \pm 3.6\%$ of the time. Our results suggest that refinements are high-quality because they incorporate specific points in the feedback.

\section{Additional Related Work}
Existing work in NLP primarily investigates using explanations for \textit{labeled outputs} to \textit{classification tasks}.
In contrast, we do not assume access to gold-labeled outputs, and we study the more general text generation setting, which classification tasks can be formulated as~\cite{Radford2019LanguageMA,raffel2020exploring,brown2020language}.
Explanations describe why a labeled output is correct, while feedback describes how to improve a candidate output.
Prior work explores ways of using explanations to train text classification models, with mixed results~\citep[][\textit{inter alia}]{camburu2018snli,stacey2021natural,pruthi2021evaluating,wiegreffe-etal-2021-measuring,hase2021can, lampinen2022can}.
A few prior works also learn from language feedback, for the purpose of ranking candidate outputs rather than generating outputs~\citep[]{weston2016dialog, li2016dialogue, hancock2019learning,li2022using}.
\citet{matiana2021cut} learn text embeddings of language feedback, where improvements could benefit the refinement-scoring step of our algorithm.

Outside of text domains, there is abundant work in reinforcement learning that leverages language in various ways~\citep[see][for an overview]{luketina2019survey}.
Prior work uses language to specify the task~\citep[``instruction following''][\textit{inter alia}]{chaplot2017gated,mahmoudieh2022zeroshot,ouyang2022training}, drive exploration~\citep{tam2022semantic}, infer reward functions~\citep[][\textit{inter alia}]{lin2022inferring, sumers2021learning, fidler2017teaching}, and train the model via strong supervision~\cite{andreas2017modular,kaplan2017beating}, reward shaping~\cite{goyal2019using}, or purely with language by providing descriptions of trajectories \citep{nguyen2021interactive}.
In contrast, we use language to correct faulty behavior.
Other work uses language feedback at test time to correct mistakes in a model's behavior, for e.g. image segmentation~\citep{rupprecht2018guide} or code generation~\cite{elgohary-etal-2020-speak,austin2021program}.
In contrast, we use feedback to \textit{train} models, and our approach does not require human intervention at test time.
\section{Conclusion}
In this work, we proposed an algorithm for training LMs to behave in line with human preferences, by learning from natural language feedback.
We validated our approach on a carefully-controlled word-removal task, showing that only large LMs (175B parameters) accurately incorporate feedback.
Using this insight, we then tested our algorithm on the real-world task of text summarization.
Our finetuning algorithm brought a GPT-3 model to roughly human-level summarization ability, using only 100 samples of human feedback.
Language feedback is a natural form of communicating with models which may make it easier for many people to provide informative, high-quality feedback.
In the long run, our work suggests many exciting avenues for future work, e.g., in guiding models with language feedback in other domains from code generation to conversational assistance.

\section{Acknowledgements}
We are grateful to Nat McAleese, Geoffrey Irving, Jeff Wu, Sam Bowman, Daniel Ziegler, Seraphina Nix, and Lennart Heim for helpful conversations and feedback.
Jérémy Scheurer and Jun Shern Chan thank Open Philanthropy for funding that enabled this research.
Ethan Perez thanks the National Science Foundation and Open Philanthropy for fellowship support.
Jon Ander Campos is supported by a doctoral grant from the Spanish MECD.
Angelica Chen and Kyunghyun Cho are supported by the NYU Center for Data Science National Science Foundation (Award 1922658).
Kyunghyun Cho is also supported by Samsung Advanced Institute of Technology (Next Generation Deep Learning: from pattern recognition to AI).
We also thank OpenAI for providing access and credits to their models via the API Academic Access Program.
\newpage

\bibliography{references}
\clearpage
\appendix
\begin{figure}[ht!]
\centering
\includegraphics[scale=0.48]{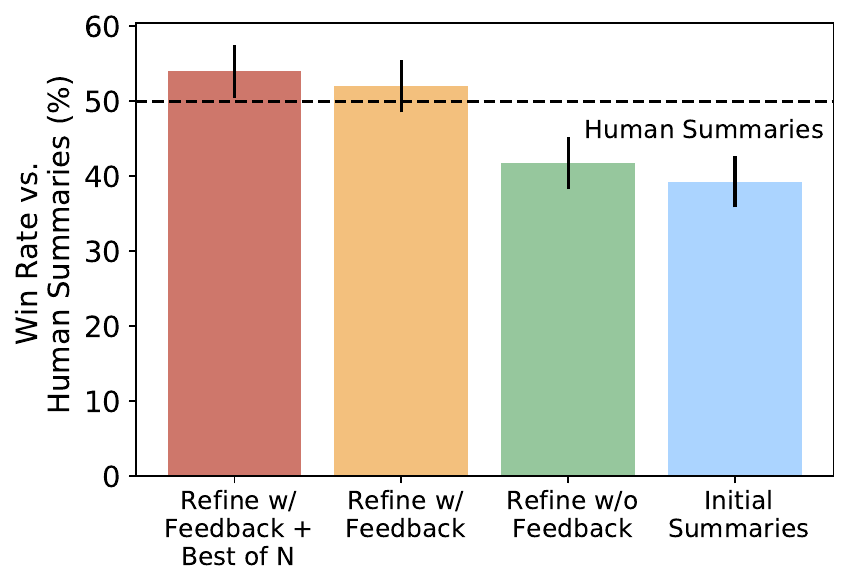}
\caption{Win rate of various refinement methods.}
\label{fig:win_rate_human_summary}
\end{figure}

\begin{figure}[ht]
\centering
\includegraphics[scale=0.48]{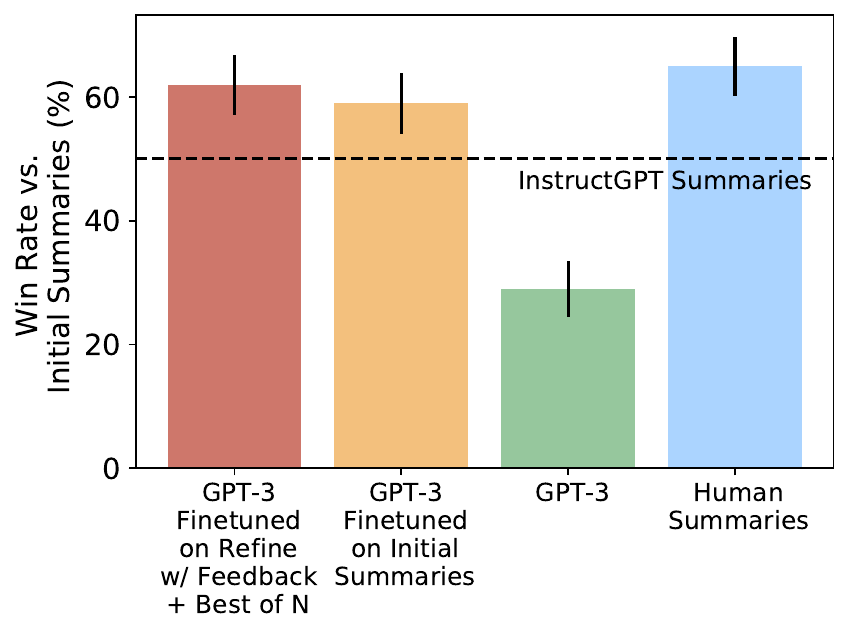}
\caption{Win rate of our learning algorithm over InstructGPT (used to generate \textsc{Initial Summaries})}.
\label{fig:finetune_win_rate_instruct_gpt}
\end{figure}


\section{Additional Results}
\label{sec:additional_results}

Here, we report additional results. In Fig.~\ref{fig:win_rate_human_summary}, we report the win rate of our various refinement methods over \textsc{Human Summaries}.
In Fig.~\ref{fig:finetune_win_rate_instruct_gpt}, we report the win rate of our learning algorithm over InstructGPT (used to generate the \textsc{Initial Summaries}).

\begin{figure}[ht]
\centering
\includegraphics[scale=0.48]{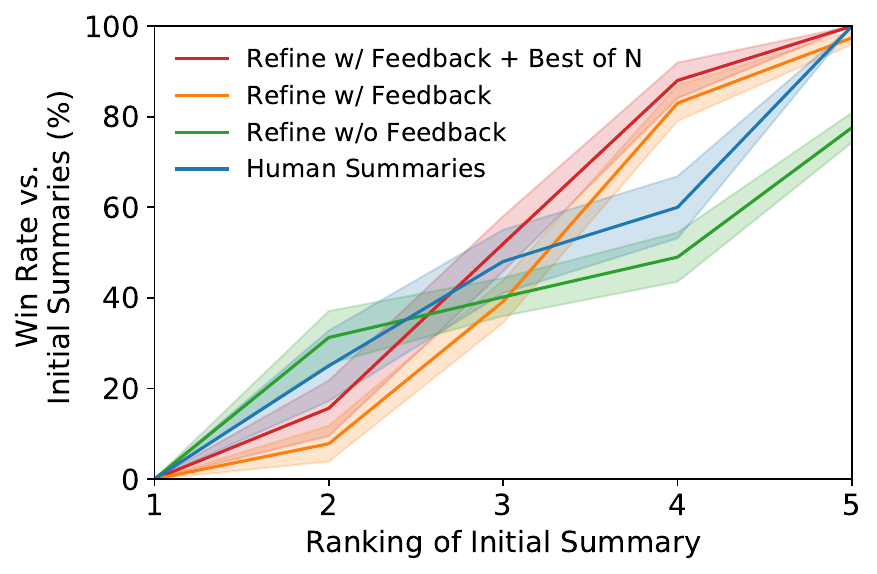}
\caption{This plot shows the win rate of various methods against the \textsc{Initial summaries} (y-axis) given various rankings of the initial summaries (x-axis). We observe that the worse the initial summaries, the better the refinements are.}
\label{fig:win_rate_across_original_summary_ranking}
\end{figure}

Fig.~\ref{fig:win_rate_across_original_summary_ranking} shows the win rate of various refinement methods by the rank of the \textsc{Initial Summaries} relative to summaries from other methods. \textsc{Refinement with Feedback (+ Best of N)} have high win rates relative to other methods when the\textsc{Initial Summary} is poorly ranked. When the \textsc{Initial Summary} rank is 4, \textsc{Refinement with Feedback} has a win rate of $83.0 \pm 3.9\%$ vs. $49.0 \pm 5.4 \%$ for \textsc{Refinement without Feedback}. On the other hand, when the \textsc{Initial Summary} is higher quality (rank 2), the win rate of \textsc{Refinement with Feedback} is only $7.8 \pm 4.0 \%$ vs. $31.25 \pm 5.8\%$ for \textsc{Refinements without Feedback}. The result is intuitive, as feedback on a bad summary should be more helpful than feedback on a good summary since there is more room for improvement on a bad summary.


\section{Targeted Word Removal Details}
\label{sec:word_removal_example}
Below is an example of how we instruct or ``prompt'' an LM to remove specific, offensive words from a sentence.
\begin{quote}
    \textit{``In this text, many toxic and offensive words are used: You are such a jerk, and a nice person, and an idiot. The ideal text should remove the word jerk, but otherwise be unchanged: You are''}
\end{quote}
Here, the target completion is \textit{`` such a nice person and an idiot.''}
More formally, we sample offensive sentences by using $k$ offensive words from a fixed set of $25$ offensive words, drawn uniformly at random (without replacement). Each offensive sentence also includes the words "nice person" in addition to all the offensive words. For each $k \in \{1, \dots, 10\}$, we sample $50$ offensive sentences. The task is then to remove $l \in [1, 2, 3]$ offensive words from a given sentence, with $k \geq l$. Since we include the words "nice person" in the offensive sentence, we can remove $l=k$ offensive words and still have a target sentence that intuitively makes sense.

\section{Human Feedback and Evaluation}
\label{sec:human_evals}
Automated metrics such as ROUGE \citep{lin-2004-rouge} do not correlate well with human preferences for summarization \citep{paulus2018deep,stiennon2020learning}, so we conduct human evaluations to evaluate various methods. We show all instructions given to annotators and evaluators \href{https://tinyurl.com/ytcz7xdm}{here}.
Two of our authors wrote feedback on the \textsc{Initial Summaries}.
To annotate feedback, they had access to the title, post, and the \textsc{Initial Summary}.
One author conducted the human evaluation for how often refinement incorporates feedback (Fig.~\ref{fig:win_rates_and_incorporating_feedback}; right). For this evaluation, we provided the title, post, \textsc{Initial Summary}, feedback, and summaries generated by various methods. Lastly, two authors not involved in feedback annotation conducted the human evaluation of generated refinements  (Fig.~\ref{fig:win_rates_and_incorporating_feedback} left, Fig.~\ref{fig:win_rate_human_summary}, and Fig.~\ref{fig:win_rate_across_original_summary_ranking}). The annotators had access to title, post, and summaries generated with various methods. See Appendix \ref{sec:ranking} for more detail about our ranking procedure. One author conducted the human evaluation for Figs.~\ref{fig:finetuning_win_rate} and~\ref{fig:finetune_win_rate_instruct_gpt}.

\section{Details about Ranking Procedure}
\label{sec:ranking}
We use a standard ranking scheme where each of $5$ summaries is given a rank between $1$ and $5$ (inclusive). The method \textsc{Refinement without Feedback} often generates refinements that are exact copies of the \textsc{Initial Summaries}, so we allow for ties in our ranking scheme. We assign the rank $r'$ to all summaries ranked in a tie, where $r'=\frac{r + (r+n-1)}{2}$, $r$ is the rank of the tied elements, and $n$ is the number of ties at the rank. For example, we map a ranking of $(1,2,2,4,5) \rightarrow (1,2.5,2.5,4,5)$ and a ranking of $(1,2,3,3,3) \rightarrow (1,2,4,4,4)$.

\section{Hyperparameters}
\label{sec:hyperparameters}
\subsection{Generating Refinements}
For the targeted word removal experiments (\S\ref{sec:targeted_word_removal}), we use greedy decoding until $200$ tokens or \textit{/\ n} is generated. For all summarization experiments (\S\ref{sec:summarization_experiments}), we sample up to $48$ tokens~\citep[as in][]{stiennon2020learning} with nucleus sampling~\cite{holtzman2019curious} with $p=0.9$. We strip non-alphanumeric characters (e.g., newlines) from the beginning of sampled summaries. Due to the maximum token length, sampled summaries sometimes end with an incomplete sentence. Thus, we remove ending sentences that do not end in ``.'', ``!'', or ``?''.

\subsection{Finetuning}
We finetune GPT-3 with 175B parameters on \textsc{Refinements with Feedback + Best of N} and \textsc{Initial Summaries}.
We use a batch size of $1$ and the default of $4$ epochs, as recommended by the OpenAI API.
For other hyperparameters, we conduct a hyperparameter search using 5-fold cross-validation on our train dataset of $100$ examples.
We use OpenAI's default hyperparameter settings from the OpenAI API and search over the \textit{learning rate multiplier}, the multiplier on the pretraining learning rate used to obtain the fine-tuning learning rate.
We sweep over $[0.005, 0.01, 0.025, 0.05, 0.1, 0.2]$ and choose $0.05$.
We also use OpenAI's default hyperparameter settings and sweep over the \textit{prompt loss weight} or the weight used for language modeling loss on the prompt \citep{Radford2018ImprovingLU}.
We sweep over $[0.01, 0.025, 0.05, 0.1, 0.2]$ and choose $0.01$.
We then finetune a new model on the full $100$ examples with the chosen hyperparameters.
We refer to the API \href{https://beta.openai.com/docs/api-reference/finetunes/create}{documentation} for a more detailed description of the adjustable hyperparameters provided by OpenAI.

\section{Prompt Templates}
\label{sec:prompts}

Table \ref{tab:prompt_template} shows the prompts used in our experiments.
\begin{table*}[ht]
    \centering
    \begin{tabular}{p{4cm} p{8cm} c}
    \toprule
    \textbf{Methods} &  \textbf{Format} \\ 
  \hline
    \textsc{Initial Summary} & Write an excellent summary of a given text. An excellent summary is coherent, accurate, concise, and detailed, and it follows human preferences about summaries.  &  \\
    & \\
    & TITLE: \{\texttt{title}\}  \\
    & \\
    & Text: \{\texttt{text}\} & \\
    & \\
    & TL;DR:  & \\ 
   \hline
 \textsc{Refinement with Feedback} & Given a text, an initial summary, and feedback on the initial summary, write an improved summary that incorporates the feedback on the initial summary and is better than the initial summary. The improved summary is coherent, accurate, concise, and detailed, and it follows human preferences about summaries. Most importantly, the improved summary incorporates the feedback.  &  \\
        & \\
        & TITLE: \{\texttt{title}\}  \\
        & \\
        & Text: \{\texttt{text}\} & \\
            & \\
        & Summary: \{\texttt{summary}\} & \\
            & \\
        & Feedback: \{\texttt{feedback}\} & \\
            & \\
        & TL;DR:  & \\ \hline
  \textsc{Refinement without Feedback} & Given a text and an initial summary, write an improved summary that is better than the initial summary. The improved summary is coherent, accurate, concise, and detailed, and it follows human preferences about summaries.  &  \\
      & \\
        & TITLE: \{\texttt{title}\}  \\
            & \\
        & Text: \{\texttt{text}\} & \\
            & \\
        & Summary: \{\texttt{summary}\} & \\
            & \\
        & TL;DR:   & \\
\bottomrule
    \end{tabular}
    \caption{Prompt templates used for various methods.}
    \label{tab:prompt_template}
\end{table*}

\section{Examples}
\label{sec:examples}
Table \ref{tab:example_summaries} shows summaries of 10 random Reddit posts from various methods: initial model-generated summaries and refinement methods.
\clearpage
\onecolumn
\begin{longtable}[h]{|p{13.5cm}|}
\hline
\textbf{\textit{Example 1}} \\
\hline
\\
\textbf{Initial Summary} \\
Should the author ask her committed boyfriend if he lied about visiting an old flame? \\
\\
\textbf{Feedback} \\
The summary is good but should mention more details. Concretely, it should remove the word "committed", since that isn't emphasized in the text. The summary should also mention that the author believes her boyfriend visited an old flame very early on in their relationship, i.e. after 4 months. Further the summary should mention that her boyfriend has already denied that this happened when the author asked him when they were both drunk. It should also convey that the boyfriend has lied before. The author is asking if she should mention this issue or not, since it would bring peace but she already had brougth it up once. \\
\\
\hdashline
\\
\textbf{Refinement with Feedback + Best of N} \\
The author is considering asking her boyfriend if he lied about visiting an old flame early on in their relationship, as this has been bothering her. He already denied it when asked drunk, and he has lied to her before. \\
\\
\textbf{Refinement with Feedback} \\
The author is considering asking her boyfriend if he lied about visiting an old flame early on in their relationship, after he denied it when asked while drunk. She is worried about how this would add to the list of his past lies. \\
\\
\textbf{Refinement without Feedback} \\
The author is considering asking her committed boyfriend if he lied about visiting an old flame, and she is torn about whether or not to do so. \\
\\
\textbf{Human Summary} \\
Asked my bf once if he went to go meet a girl, but i think he lied. Should I ask if he lied? \\
\\
\hline

\textbf{\textit{Example 2}} \\
\hline
\\
\textbf{Initial Summary} \\
my boyfriend invited me to spend Mother's Day with his parents but I feel weird about it because I don't want to imply that I will become their daughter one day. Is this too soon or am I overthinking it? \\
\\
\textbf{Feedback} \\
The summary is mostly correct but it should mention that one of the reason why the original poster feels weird about all this situation is because their relationship is just starting and they haven't talked about a future together yet. It is also important that the original poster can't spend they Mother's day with her mom and this is one of the reasons why her boyfriend has invited her. \\
\\
\hdashline
\\
\textbf{Refinement with Feedback + Best of N} \\
The original poster is feeling weird about spending Mother's Day with their boyfriend's parents because they haven't talked about a future together yet. They also can't spend Mother's Day with their own mom. \\
\\
\textbf{Refinement with Feedback} \\
The original poster feels weird about spending Mother's day with their boyfriend's parents because they haven't talked about their relationship yet, but they should go because it will be a good opportunity to get to know them better. \\
\\
\textbf{Refinement without Feedback} \\
My boyfriend invited me to spend Mother's Day with his parents, but I feel weird about it because I don't want to imply that I will become their daughter one day. Is this too soon or am I overthinking it? \\
\\
\textbf{Human Summary} \\
my boyfriend and I have been together 8 months and he invited me to spend Mother's Day with his mom and dad, but I feel uncomfortable too soon? \\
\\
\hline

\textbf{\textit{Example 3}} \\
\hline
\\
\textbf{Initial Summary} \\
Girl (F/19) doesn't know what to do with guy (M/21) who is good and polite, but not romantic and doesn't want anything serious. \\
\\
\textbf{Feedback} \\
The summary should make clear that the girl is already together with the guy for 7 months. The summary should also point out that the guy is not passionate and doesn't want sex. Lastly the summary should convey that the author is frustrated by the fact that the guy doesn't want anything serious and says he doesn't want to go fast, but that she also thinks she's in love. \\
\\
\hdashline
\\
\textbf{Refinement with Feedback + Best of N} \\
Girl (F/19) is dating guy (M/21) who is good and polite but not romantic and doesn't want anything serious. Girl is frustrated because she wants more than what the guy is willing to give. \\
\\
\textbf{Refinement with Feedback} \\
Girl (F/19) is frustrated with guy (M/21) because he is good and polite, but not romantic and doesn't want anything serious. \\
\\
\textbf{Refinement without Feedback} \\
Girl (F/19) doesn't know what to do with guy (M/21) who is good and polite, but not romantic and doesn't want anything serious. \\
\\
\textbf{Human Summary} \\
He (21) is a good guy, but I'm afraid he doesn't want anything serious with me (19).  How should I react? \\
\\
\hline

\caption{Examples of summaries, human-written feedback, and generated refinements.}
\label{tab:example_summaries}
\end{longtable}
\clearpage
\twocolumn

\end{document}